\documentclass[11pt]{article}

\usepackage[margin=1in]{geometry}
\usepackage[utf8]{inputenc}
\usepackage[T1]{fontenc}
\usepackage[colorlinks=true,linkcolor=blue!60!black,citecolor=blue!60!black,urlcolor=blue!60!black]{hyperref}
\usepackage{url}
\usepackage{booktabs}
\usepackage{amsfonts}
\usepackage{amsmath}
\usepackage{amssymb}
\usepackage{nicefrac}
\usepackage{microtype}
\usepackage{graphicx}
\usepackage{xcolor}
\usepackage{natbib}
\usepackage{caption}
\usepackage{subcaption}
\usepackage{tikz}
\usepackage{soul}
\usetikzlibrary{positioning,arrows.meta,calc}
\definecolor{vizblue}{HTML}{2A78D6}
\definecolor{vizorange}{HTML}{EB6834}
\definecolor{vizaqua}{HTML}{1BAF7A}
\definecolor{inkmuted}{HTML}{52514E}

\newif\ifprovenance
\provenancefalse
\definecolor{mwbg}{HTML}{FFE97A}
\definecolor{fixbg}{HTML}{A9DCFF}
\sethlcolor{mwbg}
\soulregister{\citep}{7}
\soulregister{\citet}{7}
\soulregister{\ref}{7}
\soulregister{\eqref}{7}
\soulregister{\url}{7}
\ifprovenance
  \DeclareRobustCommand{\mw}[1]{\hl{#1}}
  \newcommand{\mweq}[1]{\colorbox{mwbg}{$\displaystyle #1$}}
  \newcommand{\mwbox}[1]{\colorbox{mwbg}{#1}}
  \DeclareRobustCommand{\mwnote}[1]{{\color{red!65!black}\hl{[Note: #1]}}}
  \DeclareRobustCommand{\mwfix}[1]{{\sethlcolor{fixbg}\hl{#1}}}
\else
  \DeclareRobustCommand{\mwfix}[1]{#1}
  \DeclareRobustCommand{\mw}[1]{#1}
  \newcommand{\mweq}[1]{#1}
  \newcommand{\mwbox}[1]{#1}
  \DeclareRobustCommand{\mwnote}[1]{}
\fi

\graphicspath{{figs/}}

\newcommand{\FVE}{\mathrm{FVE}}
\newcommand{\ns}{\text{ns}}

\title{Analysis of Variational Sparse Autoencoders}

\author{
  Zachary Baker \\
  University of Colorado \\
  \texttt{zachary.baker@baker-research.dev}
  \and
  Yuxiao Li \\
  Beneficial AI Foundation (BAIF)
}
\date{}

\begin{document}
\maketitle
{\renewcommand{\thefootnote}{}\footnotetext{Code available at
\url{https://github.com/ZachData/SPAR-Variational-Sparse-Autoencoders}.}}

\begin{abstract}
Variational sparse autoencoders (vSAEs) replace a sparse autoencoder's
deterministic Top-K mechanism with a sampled Gaussian posterior, which is
\mw{hypothesized} to produce a more robust feature dictionary. To test this, we
build permutation comparisons over \mwfix{training seeds for} custom
analyses\mwfix{, and score two} popular benchmarks\mwfix{ on one pair of
existing checkpoints}. When fixing the unit variance and ablating
sampling on the vSAE, we find that the gap between the non-sampled vSAE and a
Top-K SAE \mw{with an $L_2$ activation penalty} can be decomposed completely
into a set of five implementation details. When we enable sampling and allow
the network to learn its own variance, we find that the discrete selection
process of Top-K fails when in conjunction with stochastic sampling methods.
At the noise scale these models are trained upon, vastly different Top-K
feature sets (Jaccard $0.075$--$0.29$, chance $0.067$) are found on equal model
activations due only to \mwfix{the sampling noise, in two forward passes of
one trained model on the same token}. Soft methods such as JumpReLU fail to
\mwfix{hold their sparsity target} as well due to sampling, though under
different failure modes. Noise moves its achieved
sparsity, and \mw{at} matched sparsity reconstruction fails to track
\mw{selection} stability ($r=-0.55$). Ablating the KL loss term recovers only
$3.9\%$ of \mw{the lost} reconstruction, the rest is due to sampling.
\end{abstract}

\section{Introduction}
\label{sec:intro}

Sparse autoencoders (SAEs) decompose language-model activations into a sparse,
overcomplete dictionary of
features~\citep{sharkey2022taking,cunningham2023sparse,bricken2023monosemanticity},
motivated by the observation that a neuron within a network can represent many
unrelated concepts via superposition~\citep{elhage2022superposition}. The
features created in this overcomplete dictionary depend on dictionary
size~\citep{bricken2023monosemanticity}. A small dictionary may represent a
larger class of a concept (general punctuation, nouns such as cars, houses,
etc.) while a large dictionary will have specific features of that concept
(commas, makes/models of cars, apartments/mansions, etc.). Thus, resolution of
concepts in features can be indirectly guided by dictionary size.

A variational method is hypothesized to enable more control. Replacing this
indirect control of dictionary size with a variational posterior as seen in a
variational autoencoder (VAE)~\citep{kingma2013autoencoding} may be a method
for direct control. The sampling noise limits how much information a single
latent can carry. Under Gaussian noise of variance $\sigma^2$, at most about
\mw{$\tfrac12\log(1+\mathrm{Var}(\mu)/\sigma^2)$} nats for \mw{a} latent
\mw{whose} mean $\mu$ varies across tokens. A vSAE has limited capacity and
must choose which distinctions minimize loss, \mw{which is} the trade-off the
\mw{$\beta$}-weighted KL of the $\beta$-VAE controls~\citep{higgins2017betavae},
and the \mw{distinctions} that separate two nearly identical latents are the
first to be washed out by noise.

We hypothesize that this sampling noise will organize the dictionary towards
more general features without adding copious amounts of computational
overhead. We test \mwfix{whether the standard construction can deliver
this} with a controlled decomposition of
\mw{existing} implementations~\citep{marks2024dictionary_learning}. Each
element of the decomposition is a \mw{permutation} test over 13 seeds \mw{per
arm}\mwfix{, enough for a single comparison to reach} statistical
significance \mwfix{of} 5 standard deviations\mwfix{ (\S\ref{sec:setup})}.

The first test run \mw{(\S\ref{sec:e1})} considers whether the vSAE with a
fixed variance (no sampling) is distinguishable from existing Top-K SAEs with
an added $L_2$ penalty on activations. The two objectives are shown to be
algebraically and experimentally identical, and we show that deviations
between the two methods can be attributed to controllable implementation
details.

Once we allow sampling on the vSAE, we analyze its selected feature dictionary
and the training run\mw{ (\S\ref{sec:mechanism})}. We test \mw{per-token} Top-K,
BatchTopK, \mw{and} JumpReLU to analyze differences\mw{
(Figure~\ref{fig:studymap})}.

\mwfix{We also measure the effect of the implementation choices: each detail
the fixed-variance comparison turns up, and one more, is tested as a
one-factor contrast. Every one moves a metric by $|d|\ge7.3$ (end of
\S\ref{sec:e1}). Appendix~\ref{sec:e4} reports a separate observation on two
downstream interpretability metrics.}

\begin{figure}[t]
\centering
\begin{tikzpicture}[
    font=\footnotesize,
    head/.style={fill=#1, text=white, font=\footnotesize\bfseries, rounded corners=2pt,
                 minimum height=5.5mm, text width=7.6cm, align=left, inner xsep=5pt, anchor=north west},
    note/.style={text width=7.75cm, align=left, inner sep=0pt, anchor=north west},
    lab/.style={font=\scriptsize, text=inkmuted},
  ]
  \def\bw{0.55}   
  \def\hw{0.17}   
  \def\hs{1.3}    
  \def\yb{-2.45}  
  \def\xe{6.95}   
  \node[head=vizblue] at (0, 0) {\S\ref*{sec:e1}\; Sampling off ($\boldsymbol\sigma=\mathbf 1$)};
  \foreach \m [count=\i from 0] in {1.0,.86,.74,.64,.56,.50,.47,.44,.38,.30,.20,.10} {
    \pgfmathsetmacro{\x}{0.4+\i*\bw}
    \ifnum\i<6
      \fill[vizblue!70] (\x-\hw, \yb) rectangle (\x+\hw, \yb+\m*\hs);
    \else
      \draw[inkmuted, line width=0.4pt] (\x-\hw, \yb) rectangle (\x+\hw, \yb+\m*\hs);
    \fi
  }
  \draw[inkmuted, line width=0.4pt] (0.1, \yb) -- (\xe, \yb);
  \draw[densely dashed, vizblue, line width=0.5pt] (0.1, \yb+0.485*\hs) -- (\xe, \yb+0.485*\hs)
     node[lab, anchor=south east, text=vizblue] {top-$k$ cutoff};
  \node[lab, anchor=north west] at (0.1, \yb-0.02) {features, sorted by their mean $\mu$};
  \node[lab, anchor=south west] at (3.75, \yb+0.9*\hs) {$\mathbf z=\boldsymbol\mu$: the same $k$ every pass};
  \node[note] at (0, -3.05) {%
    No noise reaches the selection, and
    $\mathrm{KL}=\tfrac12\|\boldsymbol\mu\|^2$: the vSAE \emph{is} a Top-K
    SAE with an $L_2$ penalty.\\[2pt]
    Before five implementation details are matched, it differs from one
    by $|d|=37$--$183$.};

  \begin{scope}[xshift=8.45cm]
  \node[head=vizorange] at (0, 0) {\S\ref*{sec:mechanism}\; Sampling on ($\boldsymbol\sigma$ learned)};
  \foreach \m [count=\i from 0] in {1.0,.86,.74,.64,.56,.50,.47,.44,.38,.30,.20,.10} {
    \pgfmathsetmacro{\x}{0.4+\i*\bw}
    \draw[inkmuted!60, line width=0.4pt] (\x-\hw, \yb) rectangle (\x+\hw, \yb+\m*\hs);
  }
  \foreach \z/\s [count=\i from 0] in {.77/1,.65/1,.59/1,.79/1,.35/0,.29/0,.39/1,.72/1,.20/0,.16/0,.24/0,.13/0} {
    \pgfmathsetmacro{\x}{0.4+\i*\bw-0.08}
    \ifnum\s=1 \fill[vizblue] (\x, \yb+\z*\hs) circle (0.7mm);
    \else \draw[vizblue, line width=0.5pt, fill=white] (\x, \yb+\z*\hs) circle (0.7mm); \fi
  }
  \foreach \z/\s [count=\i from 0] in {1.08/1,.75/1,.55/1,.37/0,.63/1,.46/0,.56/1,.45/0,.66/1,.42/0,.23/0,.11/0} {
    \pgfmathsetmacro{\x}{0.4+\i*\bw+0.08}
    \ifnum\s=1 \fill[vizorange] (\x, \yb+\z*\hs) circle (0.7mm);
    \else \draw[vizorange, line width=0.5pt, fill=white] (\x, \yb+\z*\hs) circle (0.7mm); \fi
  }
  \draw[inkmuted, line width=0.4pt] (0.1, \yb) -- (\xe, \yb);
  \node[lab, anchor=north west] at (0.1, \yb-0.02) {the same token and $\boldsymbol\mu$, two sampled passes};
  \fill[vizblue] (5.75, \yb+1.04*\hs) circle (0.7mm) node[lab, right=0.6mm] {pass 1};
  \fill[vizorange] (5.75, \yb+0.81*\hs) circle (0.7mm) node[lab, right=0.6mm] {pass 2};
  \node[lab, anchor=west] at (5.62, \yb+0.58*\hs) {filled = kept};
  \node[note] at (0, -3.05) {%
    Noise is added \emph{before} Top-$k$ and is $80$--$540\times$ the gap
    at the cutoff.\\[2pt]
    Two passes share Jaccard $0.109$ (chance $0.067$).
    Lost FVE tracks that churn on TopK and BatchTopK.};
  \end{scope}
\end{tikzpicture}
\caption{\mw{The study at a glance. The one switch is whether $\boldsymbol\sigma$ is
learned. The bars and dots are schematic, showing twelve features ($k=6$)
sorted by their mean $\mu$. Top-$k$ sees $\boldsymbol\mu$ and keeps the same features
on every pass with sampling off (\S\ref{sec:e1}), and when sampling is on
(\S\ref{sec:mechanism}), two passes over the same token keep different sets.
What separated the vSAE from a Top-K SAE with an $L_2$ penalty was five
implementation details that no stated loss shows. The end of
Section~\ref{sec:e1} switches each of them, and a ReLU on $\boldsymbol\mu$, alone, and every one is a
large effect (above $d=0.8$), at $|d|\ge7.3$ and up to $175$.}}
\label{fig:studymap}
\end{figure}

\section{Related Work}
\label{sec:related}

\mw{Sparse autoencoders for language-model interpretability were introduced by
\citet{sharkey2022taking} and \citet{cunningham2023sparse}. They build on the
superposition hypothesis of \citet{elhage2022superposition}, were scaled by
\citet{bricken2023monosemanticity}, and were moved to production-model scale
by \citet{templeton2024scaling}. Top-K sparsity is the $k$-sparse
autoencoder of \citet{makhzani2013ksparse} as
scaled to language models by
\citet{gao2024scalingevaluatingsparseautoencoders}, who also introduce the
AuxK dead-feature revival loss our models train with. BatchTopK
\citep{bussmann2024batchtopk} and JumpReLU~\citep{rajamanoharan2024jumprelu}
are the other selection mechanisms tested, and Gated
SAEs~\citep{rajamanoharan2024gated} the sparsity-penalty-side alternative.
A vSAE combines Top-K with the variational autoencoder and its
reparameterisation trick~\citep{kingma2013autoencoding}, and with the
$\beta$-weighted KL trade-off~\citep{higgins2017betavae}.
\citet{lu2025sparseautoencodersagain} also combine the two, proposing VAEase,
a hybrid of SAEs and VAEs that they report learns sparser latent
representations than SAEs and VAEs of equal capacity. JumpReLU's threshold is trained with a straight-through
estimator~\citep{bengio2013estimating}.}

\mw{Similar prior treatments of noise and discrete selection are the
Gumbel-Softmax and Concrete relaxations of discrete
sampling~\citep{jang2016categorical,maddison2016concrete}. The differentiable
perturbed optimisers of \citet{berthet2020learning} handle argmax-type
decisions such as ranking, and the hard-concrete stochastic gate of
\citet{louizos2018learning} separates a gate deciding membership from the
magnitude it passes.}

\section{Setup}
\label{sec:setup}

Our work is a modification of the existing Top-K
architecture~\citep{gao2024scalingevaluatingsparseautoencoders}, as
implemented in \texttt{dictionary\_learning}~\citep{marks2024dictionary_learning},
which encodes activations \mw{$\mathbf{x}\in\mathbb{R}^{d_{\mathrm{model}}}$ as
$\mathbf{f}=\mathrm{ReLU}(\mathbf{W}_{\mathrm{enc}}(\mathbf{x}-\mathbf{b}_{\mathrm{dec}})+\mathbf{b}_{\mathrm{enc}})$}
and keeps its largest $k$ \mwfix{latent activations}, setting the rest to 0. The
decoding step is
\mw{$\hat{\mathbf{x}}=\mathbf{W}_{\mathrm{dec}}\mathbf{f}_{\mathrm{topk}}+\mathbf{b}_{\mathrm{dec}}$}.
Sparsity is architectural in this SAE method, so \mw{the} loss carries no
$L_1$ term:
\begin{equation}
\mweq{\mathcal{L}_{\mathrm{SAE}} = \|\mathbf{x}-\hat{\mathbf{x}}\|_2^2
+ \alpha\,\mathcal{L}_{\mathrm{aux}},}
\label{eq:sae}
\end{equation}
where \mw{$\mathcal{L}_{\mathrm{aux}}$} is the AuxK loss that revives dead
features by requiring the top \mw{$k_{\mathrm{aux}}$} dead latents to explain
the residual\mw{ $\mathbf{e}=\mathbf{x}-\hat{\mathbf{x}}$. A latent counts as
dead once it has gone unselected for 1{,}000 tokens. The
$k_{\mathrm{aux}}=\min(256,\#\text{dead})$ dead latents with the largest
encoder activations are decoded on their own, without the decoder bias, and
$\mathcal{L}_{\mathrm{aux}}$ is that reconstruction's squared error against
$\mathbf{e}$, divided by the total variance of $\mathbf{e}$ over the batch,
weighted by $\alpha=1/32$. The residual is held fixed, so the term's gradient
reaches only the dead latents and the decoder columns they use.}

In our vSAE, we treat the \mwfix{latent code} as \mw{a} posterior
\mw{$q(\mathbf{z}\mid\mathbf{x})=\mathcal{N}(\boldsymbol{\mu},\boldsymbol{\sigma}^2)$}
with $\boldsymbol{\mu}$ and $\log\boldsymbol{\sigma}^2$ produced by the
encoder, and sample
\mw{$\mathbf{z}=\boldsymbol{\mu}+\boldsymbol{\epsilon}\odot\boldsymbol{\sigma}$,
$\boldsymbol{\epsilon}\sim\mathcal{N}(\mathbf{0},\mathbf{I})$}, apply Top-K to
$\mathbf{z}$, and regularize the posterior \mw{toward}
\mw{$p(\mathbf{z})=\mathcal{N}(\mathbf{0},\mathbf{I})$}:
\begin{equation}
\mweq{\mathcal{L}_{\mathrm{vSAE}} = \|\mathbf{x}-\hat{\mathbf{x}}\|_2^2
+ \beta\,\mathrm{KL}\!\left[q(\mathbf{z}\mid\mathbf{x})\,\|\,p(\mathbf{z})\right]
+ \alpha\,\mathcal{L}_{\mathrm{aux}}.}
\label{eq:vsae}
\end{equation}
This coefficient $\beta$ trades reconstruction against the \mw{strength} of
the prior in a similar method to the $\beta$-VAE found
in~\citet{higgins2017betavae}\mw{.} Noise is applied before the selection of
the \mwfix{latents}\mw{.} The encoder samples $\mathbf{z}$ and then selects
the Top-K of that $\mathbf{z}$\mw{, by magnitude $|\mathbf{z}|$
(Figure~\ref{fig:pipeline})}.

We fix the variance to isolate behaviors of the vSAE. The fixed variance KL
divergence, an $L_2$ penalty on the \mw{pre-activations}, is expressed as:
\begin{equation}
\mweq{\mathrm{KL}\!\left[\mathcal{N}(\boldsymbol{\mu},\mathbf{I})\,\|\,\mathcal{N}(\mathbf{0},\mathbf{I})\right]
= \tfrac{1}{2}\textstyle\sum_i \left(\mu_i^2 + 1 - 1 - \log 1\right)
= \tfrac{1}{2}\|\boldsymbol{\mu}\|_2^2 .}
\label{eq:kl}
\end{equation}
With the variance fixed, the encoder never samples\mw{, so}
$\mathbf{z}=\boldsymbol{\mu}$\mw{ is a fixed function of $\mathbf{x}$}. The vSAE
with fixed variance is theoretically equivalent to a Top-K SAE with an added
$L_2$ activation penalty, and we experimentally show equivalence within
float32 \mw{precision} (relative \mw{difference $\le3\times10^{-7}$ between
the two trainers' own loss functions, loaded with identical weights and
evaluated on the same batch, at initialisation and at a trained checkpoint}).

\paragraph{Setup.} We run experiments on the GELU-1L TransformerLens language
model.
\mw{All models train on residual-stream activations at layer 0
($d_{\mathrm{model}}=512$), with dictionary size $d=2048$ ($4\times$),
$k=256$, 10{,}000 steps, bfloat16 and AuxK$=1/32$ (the JumpReLU models of
\S\ref{sec:jumprelu} replace $k$ with an $L_0$ target of 256 and use their
own auxiliary loss, weight $0.1$). Training divides activations by a
normalising constant ($\approx25.5$ at this hook point) and defines the noise
scale $\sigma$ in that normalised space; every sampled model is evaluated
there, in batches of 1{,}024 (the training batch size).}

\begin{figure}[t]
\centering
\begin{tikzpicture}[
    font=\small,
    >={Stealth[length=2.2mm]},
    box/.style={draw=inkmuted, rounded corners=2pt, fill=white, minimum height=10mm,
                inner sep=3pt, align=center},
    enc/.style={box, minimum width=3.7cm},
    sel/.style={box, fill=vizblue!12, draw=vizblue, minimum width=1.4cm},
    dec/.style={box, minimum width=3.2cm},
    noise/.style={box, fill=vizorange!14, draw=vizorange, minimum width=2.6cm},
    arr/.style={->, draw=inkmuted, line width=0.6pt},
    lab/.style={font=\footnotesize\itshape, text=inkmuted},
    loss/.style={font=\footnotesize, anchor=west, align=left},
    call/.style={circle, fill=black, text=white, font=\scriptsize\bfseries,
                 inner sep=1pt, minimum size=4mm},
    note/.style={font=\footnotesize, align=left, text width=#1},
  ]
  \def\xe{2.3}\def\xs{5.75}\def\xk{8.3}\def\xd{10.9}\def\xh{12.9}
  \def\ya{0}\def\yb{-2.0}

  \node[lab, anchor=west] at (-0.6, \ya+0.85) {Top-K SAE (the baseline)};
  \node (x1) at (-0.6, \ya) {$\mathbf{x}$};
  \node[enc] (e1) at (\xe, \ya) {encoder\\{\footnotesize$\mathrm{ReLU}(\mathbf{W}_{\mathrm{enc}}(\mathbf{x}-\mathbf{b}_{\mathrm{dec}})+\mathbf{b}_{\mathrm{enc}})$}};
  \node[sel] (k1) at (\xk, \ya) {Top-$k$\\of $\mathbf{f}$};
  \node[dec] (d1) at (\xd, \ya) {decoder\\$\mathbf{W}_{\mathrm{dec}}\mathbf{f}_{\mathrm{topk}}+\mathbf{b}_{\mathrm{dec}}$};
  \node (y1) at (\xh, \ya) {$\hat{\mathbf{x}}$};
  \draw[arr] (x1) -- (e1);
  \draw[arr] (e1) -- node[above, font=\footnotesize] {$\mathbf{f}$}
                     node[below, lab] {no sampling} (k1);
  \draw[arr] (k1) -- (d1);
  \draw[arr] (d1) -- (y1);
  \node[loss] at ($(y1.east)+(1mm,0)$) {$\|\mathbf{x}-\hat{\mathbf{x}}\|^2$\\$+\,\alpha\mathcal{L}_{\mathrm{aux}}$};

  \node[lab, anchor=west] at (-0.6, \yb+0.85) {variational SAE};
  \node (x2) at (0, \yb) {$\mathbf{x}$};
  \node[enc] (e2) at (\xe, \yb) {encoder\\$\boldsymbol{\mu}$, \;$\log\boldsymbol{\sigma}^2$};
  \node[noise] (r2) at (\xs, \yb) {sample\\$\mathbf{z}=\boldsymbol{\mu}+\boldsymbol{\epsilon}\odot\boldsymbol{\sigma}$};
  \node[sel] (k2) at (\xk, \yb) {Top-$k$\\of $|\mathbf{z}|$};
  \node[dec] (d2) at (\xd, \yb) {decoder\\$\mathbf{W}_{\mathrm{dec}}\mathbf{z}_{\mathrm{topk}}+\mathbf{b}_{\mathrm{dec}}$};
  \node (y2) at (\xh, \yb) {$\hat{\mathbf{x}}$};
  \draw[arr] (x2) -- (e2);
  \draw[arr] (e2) -- (r2);
  \draw[arr, draw=vizorange, line width=1pt] (r2) -- (k2);
  \draw[arr] (k2) -- (d2);
  \draw[arr] (d2) -- (y2);
  \node[loss] (loss2) at ($(y2.east)+(1mm,0)$)
       {$\|\mathbf{x}-\hat{\mathbf{x}}\|^2$\\$+\,\beta\,\mathrm{KL}+\alpha\mathcal{L}_{\mathrm{aux}}$};
  \node[font=\footnotesize, text=inkmuted, below=0.5mm of r2] {$\boldsymbol{\epsilon}\sim\mathcal{N}(\mathbf{0},\mathbf{I})$};

  \foreach \c in {e, k, d} {
    \draw[inkmuted!35, densely dotted, line width=0.4pt] (\c1.south) -- (\c2.north);
  }

  \node[call] (c1) at ($(loss2.north)+(0mm,2mm)$) {1};
  \node[call] (c2) at ($(r2.east)!0.5!(k2.west)+(0,6.5mm)$) {2};
  \node[note=7.4cm, anchor=north west] (n1) at (-0.6,\yb-1.15)
    {\textbf{(1) \S\ref{sec:e1}.} Public implementations default to
     $\boldsymbol{\sigma}=\mathbf{1}$. Then $\mathrm{KL}=\tfrac12\|\boldsymbol{\mu}\|^2$:
     the vSAE \emph{is} a Top-K SAE with an $L_2$ penalty.};
  \node[note=7.4cm, anchor=north west] (n2) at ($(n1.north east)+(3mm,0)$)
    {\textbf{(2) \S\ref{sec:mechanism}.} With
     $\boldsymbol{\sigma}$ free, noise is added \emph{before} Top-$k$, so it changes
     \emph{which} features are kept. Measured on three selection rules.};
\end{tikzpicture}
\caption{\mw{A Top-K SAE (top) and a variational SAE (bottom) as implemented. The
vSAE adds a variance head, a sampling step (orange) and a KL term. The sampling
step comes \emph{before} the Top-$k$ selection (blue), so noise can change which
features are selected, not just their values. The numbered markers show where
each part of this paper intervenes, (1) the KL term at fixed variance and
(2) the noise-before-selection step.}}
\label{fig:pipeline}
\end{figure}

\begin{figure}[t]
\centering
\includegraphics[width=\textwidth]{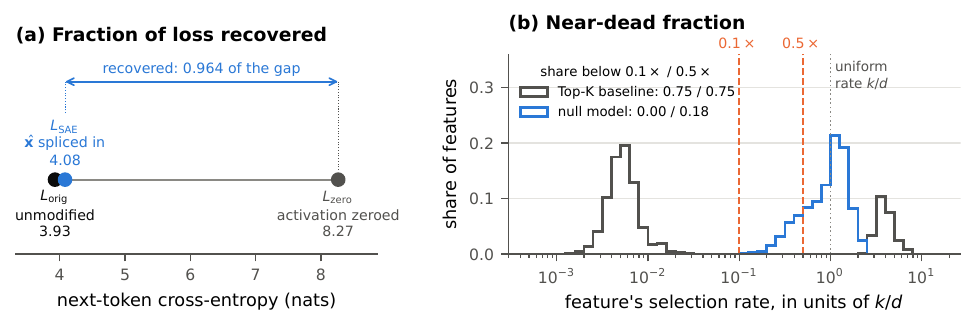}
\caption{\mw{Two of the metrics on real models. (a)~Fraction of loss recovered,
on the Top-K baseline (13-seed means). Splicing the reconstruction in costs
$L_{\mathrm{SAE}}-L_{\mathrm{orig}}$, against
$L_{\mathrm{zero}}-L_{\mathrm{orig}}$ for deleting the activation altogether.
(b)~Every feature's selection rate in units of the uniform rate $k/d$ for the
Top-K baseline and the Top-K\,+\,$L_2$ null model of \S\ref{sec:e1}. The two
dashed lines are the near-dead thresholds. The baseline's usage splits into a
rarely used majority and a heavily used minority. The null model's usage sits
around the uniform rate.}}
\label{fig:metrics}
\end{figure}

\paragraph{Terms.}\mw{ FVE, fraction of loss recovered and the two liveness
thresholds are the four metrics.}
\begin{itemize}
\item \emph{The fraction of \mw{variance} explained}\mw{ (FVE): the share of
  the activations' variance that the reconstruction reproduces,}
  \[
  \mweq{\FVE = 1-\frac{\sum_j\mathrm{Var}(x_j-\hat{x}_j)}{\sum_j\mathrm{Var}(x_j)},}
  \]
  \mw{with $j$ running over the $512$ activation dimensions and each variance
  taken over the tokens of an evaluation batch. $1$ is a perfect
  reconstruction and $0$ is no better than predicting the mean activation.}
\item \emph{Fraction of loss recovered}\mw{ (frac.~rec.): the same question
  asked of the language model's next-token cross-entropy. Splicing
  $\hat{\mathbf{x}}$ into the forward pass in place of $\mathbf{x}$ gives a
  loss $L_{\mathrm{SAE}}$. $L_{\mathrm{orig}}$ is the unmodified model's loss
  and $L_{\mathrm{zero}}$ the loss with the activation replaced by zeros:}
  \[
  \mweq{\text{frac.~rec.} = \frac{L_{\mathrm{zero}}-L_{\mathrm{SAE}}}{L_{\mathrm{zero}}-L_{\mathrm{orig}}},}
  \]
  \mw{the share of the damage done by deleting the activation that the
  reconstruction undoes (Figure~\ref{fig:metrics}a).}
\item \emph{The \mw{Jaccard} selection}\mw{ (sampled models, from
  \S\ref{sec:mechanism} on; descriptive). Run the same token through the
  encoder twice, with independent noise, and let $S_1,S_2$ be the two sets of
  $k$ selected features:}
  \[
  \mweq{J = \frac{|S_1\cap S_2|}{|S_1\cup S_2|},}
  \]
  \mw{averaged over tokens. $J=1$ when the noise never changes which features
  are selected. Two unrelated random choices of $k$ of the $d$ features share
  $k^2/d$ features on average, so chance is $J=k/(2d-k)=0.067$.
  Figure~\ref{fig:churn}a shows one token.}
\item \emph{Liveness}\mw{, measured as the near-dead fraction at two
  thresholds. Top-K selects $k$ of $d$ features per token, so a feature used
  at the uniform rate is selected on a fraction $k/d=0.125$ of tokens. With
  $u_i$ the fraction of $10^6$ tokens on which feature $i$ is selected, we
  report}
  \[
  \mweq{\tfrac1d\,\#\{\,i : u_i < r\cdot k/d\,\}\qquad\text{for } r=0.1 \text{ and } r=0.5,}
  \]
  \mw{where $\#$ counts the elements of a set (here, the features below the
  threshold), so dividing by $d$ gives the share of the dictionary used at
  under a tenth and under half of the uniform rate. This is invisible to the
  share of features selected at least once. Over $10^6$ tokens nearly every
  feature is selected sometimes, and the Top-K baseline scores $0.993$ on it
  while $0.754$ of its features sit below the tighter threshold
  (Figure~\ref{fig:metrics}b). The baseline's usage is bimodal and unaffected
  by the choice of threshold, while the null model of \S\ref{sec:e1} has
  almost no features below $0.1\times$ but $0.184$ below $0.5\times$. A
  liveness result counts only if both thresholds move in the same
  direction.}
\item \mw{\emph{Effect size}, Cohen's $d$. For two arms $a$ and $b$, with
  means $\bar a,\bar b$ and standard deviations $s_a,s_b$ over their seeds,}
  \[
  \mweq{d = \frac{\bar a-\bar b}{\tfrac12(s_a+s_b)},}
  \]
  \mw{the difference in means measured in units of the seed-to-seed spread.
  At $d=1$ the two arms' seeds overlap heavily; by $d\approx4$, 13 seeds per
  arm typically no longer overlap at all. Seeds vary little here (the SD of FVE
  across seeds is about $0.001$), so small absolute differences give large
  $d$. For example, rung R1's $d=16.5$ on FVE in \S\ref{sec:e1} is FVE $0.828$ against the
  null model's $0.846$. $d$ says how cleanly two arms separate, not how much
  the difference matters, so we give the difference itself where it matters.}
\end{itemize}

\paragraph{Design.} To guarantee that we are not measuring outliers\mw{,} we
analyze many training seeds to guarantee statistical significance. Each
training configuration, which we define as an arm, consists of \mw{13
training runs identical except for the random seed; two arms are compared by
a} two-sided permutation test over \mw{their} 13 seeds\mw{ each}.

\begin{figure}[t]
\centering
\begin{tikzpicture}[
    font=\footnotesize,
    >={Stealth[length=1.8mm]},
    arm/.style={draw=#1, fill=#1!10, rounded corners=2pt, minimum width=2.1cm,
                minimum height=9mm, align=center},
    stepbox/.style={draw=inkmuted, rounded corners=2pt, text width=4.75cm, align=left,
                 inner sep=4pt, anchor=north},
    num/.style={circle, fill=black, text=white, font=\scriptsize\bfseries,
                inner sep=0.5pt, minimum size=3.8mm},
    arr/.style={->, draw=inkmuted, line width=0.6pt},
    lab/.style={font=\scriptsize, text=inkmuted, align=center},
  ]
  \foreach \y/\c/\name/\m/\tip in {0/inkmuted/{null model}/a/10.8, -1.25/vizblue/{rung R1}/b/6.8} {
    \node[arm=\c] (arm\m) at (1.05, \y) {\textbf{\name}};
    \draw[inkmuted!45, line width=0.4pt] (6.6, \y) -- (15.0, \y);
    \draw[arr, line width=0.7pt] (arm\m.east) -- node[pos=0.22, above, lab] {train 13 seeds,}
                                    node[pos=0.22, below, lab] {evaluate each} (\tip, \y);
  }
  \foreach \x in {11.0,11.35,11.6,11.8,11.95,12.1,12.22,12.35,12.5,12.65,12.85,13.1,13.5} {
    \fill[inkmuted] (\x, 0) circle (0.6mm);
  }
  \foreach \x in {7.0,7.35,7.6,7.8,7.95,8.1,8.22,8.35,8.5,8.65,8.85,9.1,9.5} {
    \fill[vizblue] (\x, -1.25) circle (0.6mm);
  }
  \draw[black, line width=1.1pt] (12.25, -0.2) -- (12.25, 0.2) node[above, font=\scriptsize] {mean $\bar a$};
  \draw[vizblue, line width=1.1pt] (8.25, -1.05) -- (8.25, -1.45) node[below, font=\scriptsize] {mean $\bar b$};
  \draw[<->, draw=black, line width=0.5pt] (8.25, -0.62) -- node[above, font=\scriptsize] {$\Delta$} (12.25, -0.62);
  \draw[densely dotted, vizblue] (8.25, -1.05) -- (8.25, -0.62);
  \draw[densely dotted, inkmuted] (12.25, -0.2) -- (12.25, -0.62);
  \node[lab, anchor=west] (dl) at (13.65, 0.42) {one dot per seed};
  \draw[inkmuted, line width=0.4pt] (dl.west) -- (13.55, 0.07);

  \node[num] at (0.05, 0.95) {1};
  \node[anchor=west, font=\footnotesize\bfseries] at (0.3, 0.95) {An arm};
  \node[lab] at (10.3, 0.95) {each run gives one value of each metric};

  \node[stepbox] (s2) at (2.45, -2.25) {%
    \hspace*{3mm}\textbf{Difference and effect size.}\\[2pt]
    $\Delta=\bar a-\bar b$, the difference in arm means.\\[2pt]
    $d=\Delta\,/\,\tfrac12(s_a+s_b)$, with $s$ the standard deviation across an
    arm's 13 seeds. $d$ measures how cleanly the seeds separate, not how big
    $\Delta$ is.};
  \node[stepbox] (s3) at (7.75, -2.25) {%
    \hspace*{3mm}\textbf{Permutation $p$.}\\[2pt]
    Pool the 26 values, deal them at random into two groups of 13, recompute
    the difference $\Delta^{*}$. Repeat $4\times10^{6}$ times.\\[2pt]
    $p=\dfrac{\#\{|\Delta^{*}|\ge|\Delta|\}+1}{4\times10^{6}+1}$,\\[2pt]
    where $\#$ is the number of shuffles that reach $|\Delta|$.};
  \node[stepbox] (s4) at (13.05, -2.25) {%
    \hspace*{3mm}\textbf{Reported in $\sigma$.}\\[2pt]
    $p=2\,(1-\Phi(\sigma))$, so $5\sigma$ is $p\approx5.7\times10^{-7}$.\\[2pt]
    If no shuffle reaches $\Delta$, $p=2.5\times10^{-7}$, or $5.16\sigma$:
    the design's floor, whatever the effect size.};
  \foreach \k/\n in {s2/2, s3/3, s4/4} {
    \node[num] at ($(\k.north west)+(0.02,-0.02)$) {\n};
  }
  \draw[arr] (s2.east) -- (s3.west);
  \draw[arr] (s3.east) -- (s4.west);
  \draw[arr] (11.3, -1.25) -- (11.3, -1.85) -| (s2.north);
\end{tikzpicture}

\vspace{3mm}
\includegraphics[width=\textwidth]{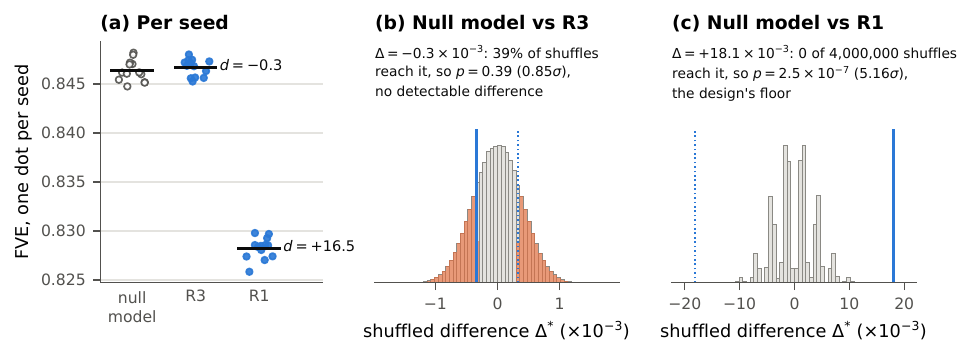}
\caption{\mw{The statistical design. Top: the four steps of every comparison
(the dots are schematic). Bottom: the same steps on two real comparisons from
\S\ref{sec:e1}, FVE of the null model against rungs R3 and R1.
(a)~Each arm's 13 seeds, with the arm mean (bar) and Cohen's $d$ against the
null model. (b,~c)~The differences produced by shuffling the 26 values
($2\times10^5$ shuffles drawn for display); the solid line is the observed
difference $\Delta$, the dotted line its mirror image. For R3, $39\%$ of
shuffles are at least as extreme (orange), so there is no detectable
difference. For R1, $\Delta$ lies beyond every shuffle, and $p$ is the
design's floor, which a larger effect could not lower.}}
\label{fig:design}
\end{figure}

\mw{We analyze each training run as one data point, and every comparison
follows the same four steps (Figure~\ref{fig:design}).}
\begin{enumerate}
\item \mw{\emph{An arm is one training configuration, trained 13 times.} The
  13 runs are identical except for the random seed, which also sets how
  activations are drawn into batches and, if the model is sampled, the
  noise. Each run gives one value of each metric, so an arm is 13 numbers per
  metric.}
\item \mw{\emph{Compare the means.} $\Delta$ is the difference between the
  two arms' means, and Cohen's $d$ (above) is $\Delta$ in units of the
  seed-to-seed spread.}
\item \mw{\emph{Shuffle to get $p$.} Pool the 26 values, deal them at random
  into two groups of 13, and recompute the difference; $p$ is the fraction of
  shuffles whose difference is at least as large in magnitude as $\Delta$
  (two-sided). We sample $4\times10^6$ shuffles, enough for $p$ to reach
  $5\sigma$.}
\item \mw{\emph{Report $p$ in $\sigma$.} $p=2\,(1-\Phi(\sigma))$, the
  two-sided Gaussian tail, so $5\sigma$ is $p\approx5.7\times10^{-7}$.}
\end{enumerate}
\mw{The test has a floor that does not depend on the effect. However far apart
two arms are, the best possible outcome is that no shuffle matches $\Delta$,
which gives $p=1/(4\times10^6+1)=2.5\times10^{-7}$, or $5.16\sigma$.
Enumerating every split would lower this only to
$2/\binom{26}{13}=1.9\times10^{-7}$ ($5.21\sigma$). With $n$ seeds per arm
the floor is $2/\binom{2n}{n}$, and 13 is the first $n$ at which it clears
$5\sigma$.}

\mwfix{We follow the falsification framework of
POPPER~\citep{huang2025popper}, which tests a broad claim through narrower
comparisons that meet two conditions. First, the broad claim's null must imply
each comparison's null, which for an implementation factor holds only when
the two arms differ in that factor alone, so every comparison here is a
one-factor contrast; where an uncontrolled variable moves with the factor
(achieved sparsity in \S\ref{sec:jumprelu}, live dictionary size in
Appendix~\ref{sec:e4}), we match it first. Second, the data used to choose a
test must not be the data it is run on. The metrics and main comparisons were
pre-registered, the KL coefficient for sampled models was chosen on
single-seed pilot runs separate from the 13 test seeds, and comparisons
designed after earlier results were seen are labelled exploratory.}

\section{The KL Loss Term}
\label{sec:e1}

\mw{We compare the released vSAE trainer ($\beta=1$, fixed variance) against a
Top-K SAE whose loss adds $\tfrac{1}{2}\lambda\|\mathbf{f}\|_2^2$ with
$\lambda=\beta$. With a 1{,}000-step KL warmup and no pre-encoder bias, the
vSAE differs from this null model by $|d|=51$ on FVE, $37$ on frac.~rec.~and
$57$--$183$ on liveness (six seeds against 13, at that design's $4.13\sigma$
ceiling).}

\begin{figure}[t]
\centering
\includegraphics[width=\textwidth]{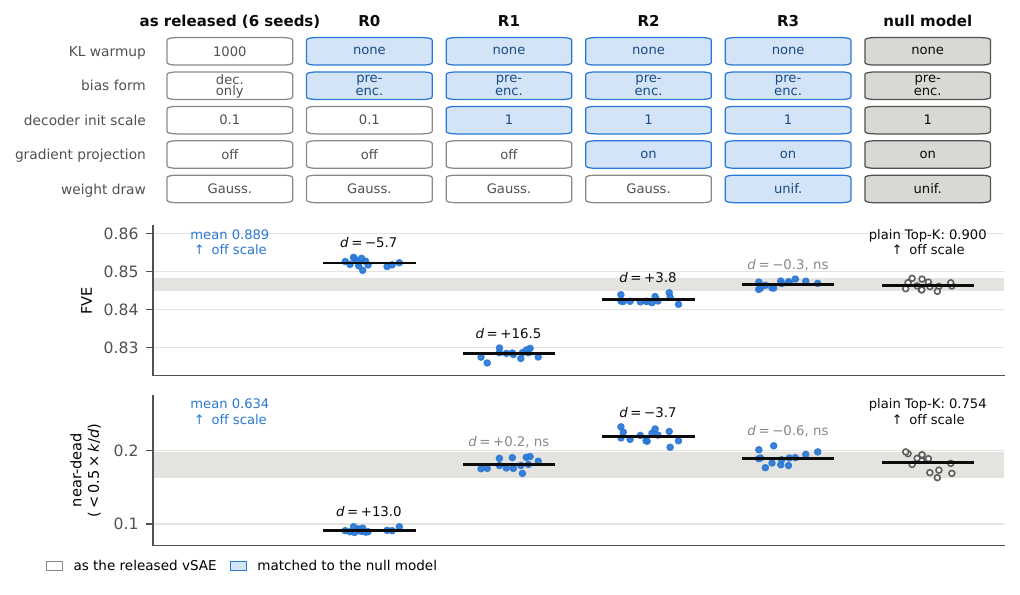}
\caption{\mw{The decomposition ladder. Top: what each arm uses for each of the
five implementation details; blue cells are matched to the null model, white
cells are as the released vSAE has them, and each rung matches one more
detail than the last. Below: FVE and the looser near-dead fraction per seed
(dots; 13 per arm), the arm mean (bar), Cohen's $d$ against the null model
(ns: $p>0.05$), and the null model's range over seeds (grey band). For reference, the plain
Top-K SAE with no penalty scores FVE $0.900$ and near-dead fraction $0.754$,
off the scale of both panels. R1 closes
the liveness gap and opens a larger FVE gap; R2 closes most of the FVE gap
and pushes liveness out of the band the other way; only R3 sits inside the
band on both. All four metrics are in Table~\ref{tab:e1ladder}
(Appendix~\ref{app:ladder}).}}
\label{fig:e1ladder}
\end{figure}

\mw{The ladder matches five implementation details cumulatively, the KL
warmup, the bias
parameterisation, the decoder's initial scale, a projection of the decoder
gradient, and the distribution the initial decoder weights are drawn from
(Figure~\ref{fig:e1ladder}; Cohen's $d$ on all four metrics in
Table~\ref{tab:e1ladder}, Appendix~\ref{app:ladder}). R0 matches the two
configuration-level differences, warmup and bias form. Its liveness differs in
opposite directions at the two thresholds ($-2.8$ and $+13.0$), a change in the shape
of the usage distribution rather than its location. Matching the decoder's
initial scale (R1) closes the liveness gap and opens a larger reconstruction
gap ($d=16.5$). Matching the decoder-gradient projection (R2) closes
$78.9\%$ of the FVE gap and $74.9\%$ of the frac.~rec.~gap but moves
liveness \emph{away from} the null model ($0.1816\to0.2197$, against
$0.1836$). The last factor is the initial weight draw. The null model draws
decoder columns as normalised uniform vectors (the default initialisation of
a linear layer), the vSAE as normalised Gaussian ones.}

\begin{table}[t]
\centering
\caption{\mw{Six one-line implementation details, each a one-factor contrast
between two arms that differ in it alone, with $|d|$ on the four metrics and the range of significance across them. Every entry above the rule
has $p\le0.05$. The first two details were matched by configuration before
the 13-seed battery and are measured on the six-seed arms of the discarded
pilot generation; their $\sigma$ is that design's ceiling, not a property of
the effect. $^\dagger$The two liveness
thresholds disagree in direction, a shape change. Below the rule is the
gradient projection switched off in a penalty-free Top-K SAE, null on every
metric ($p\ge0.22$; exploratory).}}
\label{tab:variance}
\small\setlength{\tabcolsep}{4pt}
\mwbox{%
\begin{tabular}{lcccccc}
\toprule
Detail & seeds & FVE & frac.~rec. & $<0.1\times$ & $<0.5\times$ & $\sigma$ \\
\midrule
Bias form (pre-encoder bias or not) & 6 v 13 & $39.7$ & $31.7$ & $121.7$ & $174.5$ & $4.13$ \\
KL warmup (1{,}000 steps or none) & 6 v 6 & $6.9$ & $2.9$ & $8.6$ & $2.9$ & $3.07$ \\
Initial decoder scale (0.1 or 1) & 13 v 13 & $23.1$ & $30.3$ & $2.8^\dagger$ & $18.0^\dagger$ & $5.03$--$5.16$ \\
Decoder-gradient projection & 13 v 13 & $14.3$ & $15.3$ & $2.4$ & $5.0$ & $4.48$--$5.16$ \\
Initial weight distribution & 13 v 13 & $4.7$ & $7.3$ & $2.2$ & $3.6$ & $4.31$--$5.16$ \\
ReLU on $\boldsymbol{\mu}$ & 13 v 13 & $32.5$ & $24.0$ & $19.3$ & $15.3$ & $5.03$--$5.16$ \\
\midrule
Projection, penalty-free Top-K & 13 v 13 & $0.3$ & $0.5$ & $0.3$ & $0.2$ & \ns \\
\bottomrule
\end{tabular}}
\end{table}

\mw{Matching every implementation detail removes any measurable difference
between the vSAE and the null model. R3 differs
from the null model by $-0.0003$ in FVE ($d=-0.3$, $p=0.39$) and $-0.0004$ in
frac.~rec.~($d=-0.8$, $p=0.07$), and neither liveness threshold is
significant, at the power that detected every earlier rung at $5\sigma$. This
is a failure to detect a difference, not a formal equivalence test, but the
residuals (0.0003 FVE, 0.006 in the near-dead fraction) are two orders of
magnitude smaller than the gaps that opened and closed along the ladder.}

\mw{Table~\ref{tab:variance} switches each of the five details alone, plus a
sixth, a ReLU the released vSAE applies to $\boldsymbol\mu$ before sampling.
Every one moves a metric by $|d|\ge7.3$.}

\section{\texorpdfstring{\mw{The Sampled vSAE}}{The Sampled vSAE}}
\label{sec:mechanism}

\subsection{\texorpdfstring{\mw{Sampling Destroys the Reconstruction}}{Sampling Destroys the Reconstruction}}

\begin{figure}[t]
\centering
\includegraphics[width=\textwidth]{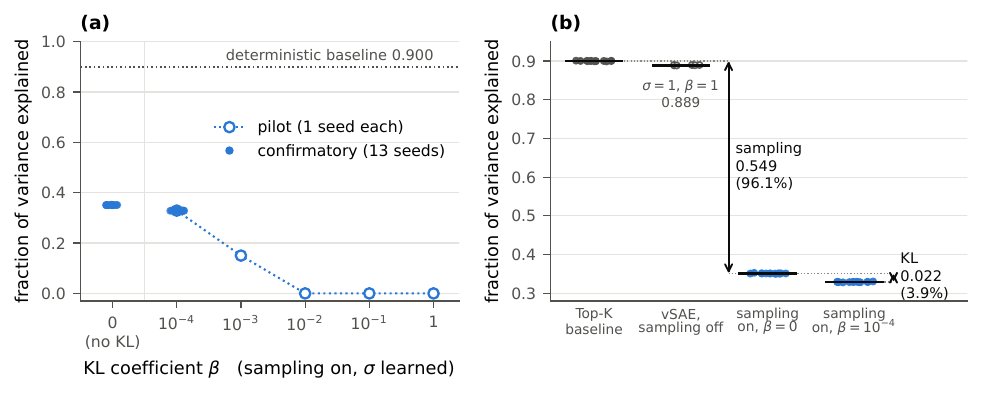}
\caption{\mw{Sampling destroys the reconstruction. (a)~This shows FVE at every
KL coefficient tested with sampling on (open circles are single-seed runs and
dots are the seeds of the two 13-seed arms). Even at $\beta=0$, where the KL contributes
nothing to the loss, FVE stays near 40\% of the deterministic baseline
(dotted). (b)~Per-seed values for the deterministic Top-K baseline, the same vSAE code
the sampled arms use with sampling off ($\boldsymbol\sigma=\mathbf 1$, the
released configuration, at $\beta=1$), and the two 13-seed sampled
arms. Of the 0.571 FVE gap between the baseline and $\beta=10^{-4}$, removing
the KL recovers 0.022.}}
\label{fig:e2split}
\end{figure}

\mw{At fixed variance the model never samples. With sampling on ($\sigma$
learned), no KL coefficient in $\{10^{-4},\ldots,1\}$ recovers a healthy
model. FVE falls monotonically from $0.329$ at $\beta=10^{-4}$ to zero by
$\beta=10^{-2}$, against the deterministic baseline's $0.900$
(Figure~\ref{fig:e2split}a). The 13-seed arm sits at the coefficient
selected from single-seed pilots (\S\ref{sec:setup}). With $\beta=0$, so that
the KL contributes nothing to the loss, FVE rises only to $0.351$. The same code
with sampling off ($\boldsymbol\sigma$ fixed at $\mathbf 1$, as released) loses
almost nothing even at the full $\beta=1$, with FVE $0.889$ (six seeds of the pilot
generation). Of the $0.571$ gap between
the baseline and $\beta=10^{-4}$, removing the KL recovers $3.9\%$. The other
$96.1\%$ is the cost of sampling itself (Figure~\ref{fig:e2split}b; 13 seeds
per arm, SD $\le0.0007$).}

\subsection{\texorpdfstring{\mw{Noise and the Selected Set}}{Noise and the Selected Set}}

\begin{figure}[t]
\centering
\includegraphics[width=\textwidth]{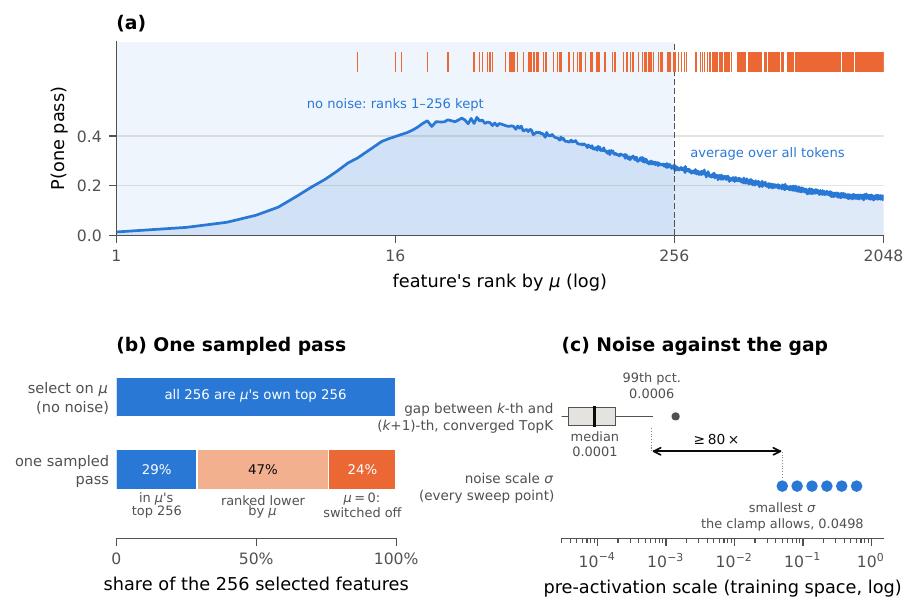}
\caption{\mw{What noise before Top-$k$ does to the selected set. One trained TopK vSAE ($\log\sigma^2_{\mathrm{init}}=-3$, one seed) in the regime it was trained in, 8{,}192 tokens. (a)~For each rank by $\boldsymbol\mu$, the probability that the feature there is selected by exactly one of two independent sampled passes (shaded ranks are kept without noise). Above the curve, the 392 features one pass keeps and the other drops for one token (Jaccard $0.13$). (b)~Of the 256 features one sampled pass selects, averaged over tokens, $29\%$ are in $\boldsymbol\mu$'s own top 256, $47\%$ rank lower, and $24\%$ have $\mu=0$, switched off by the encoder's ReLU and selected on noise alone. (c)~The gap between the $k$-th and $(k{+}1)$-th pre-activation on a converged deterministic TopK SAE, against every $\sigma$ the sweep reached. The box spans the 25th to 75th percentile, the whiskers the 5th to 99th, and the dot is the maximum.}}
\label{fig:churn}
\end{figure}

\mw{Top-K selection is an argmax over pre-activations where noise changes the
selected set whenever it exceeds the gap between the $k$-th and $(k{+}1)$-th largest
pre-activation. On a converged deterministic checkpoint that gap has median
$0.0001$ and 99th percentile $0.0006$ (Figure~\ref{fig:churn}c). The smallest
noise the reparameterisation clamp allows, $\sigma=0.0498$, is two to three
orders of magnitude larger ($80$--$540\times$), so at every noise scale this
architecture allows, the noise scrambles the selection.}

\mw{The noise is large against $\boldsymbol{\mu}$ as well. In one trained
checkpoint ($\log\sigma^2_{\mathrm{init}}=-3$), $71\%$ of the 256 features a
sampled forward pass selects lie outside $\boldsymbol{\mu}$'s own top 256, and
$24\%$ have $\mu=0$ (Figure~\ref{fig:churn}b). Two passes on the same token share a mean Jaccard
of $0.109$ against a chance level of $0.067$, and the churn reaches the top of
the ranking, not only the $k$-th place (Figure~\ref{fig:churn}a).}

\subsection{\texorpdfstring{\mw{Under Hard-$k$ Selection}}{Under Hard-k Selection}}
\label{sec:sweep}

\mw{We swept the initial noise scale over
$\log\sigma^2_{\mathrm{init}}\in\{-1,\allowbreak-2,\allowbreak-3,\allowbreak-4,\allowbreak-5,\allowbreak-8\}$
(the last clamped to an effective $-6$) on two hard-$k$ architectures, the
per-token TopK vSAE and a BatchTopK vSAE~\citep{bussmann2024batchtopk} that
selects the top $k\cdot B$ activations across a batch of $B$ tokens rather
than exactly $k$ per token, so that one token's noise-promoted feature can be
offset by another's demoted one. At each point we measure FVE and selection
Jaccard (\S\ref{sec:setup}), the latter over 8{,}192 tokens
(Table~\ref{tab:sweep}). The sweep was
designed after \S\ref{sec:mechanism}'s first results and is exploratory, as
is the JumpReLU grid of \S\ref{sec:jumprelu}.}

\begin{table}[t]
\centering
\caption{\mw{The noise-scale sweep, with FVE, selection Jaccard and, for JumpReLU,
achieved $L_0$. 13 seeds per point except the four JumpReLU points marked
$^\ast$ (5 seeds). Across seeds, SD $\le0.0026$ for FVE, $\le0.0102$ for
Jaccard and $\le5.9$ for $L_0$. Deterministic baselines are FVE $0.9002$ (TopK),
$0.9508$ (BatchTopK), $0.9213$ (JumpReLU, $L_0=269.5$). Chance Jaccard is
$0.067$. JumpReLU rows from $-3.5$ down form the $L_0$-matched band of
\S\ref{sec:jumprelu}.}}
\label{tab:sweep}
\small\setlength{\tabcolsep}{4.5pt}
\mwbox{%
\begin{tabular}{rcccccccc}
\toprule
& & \multicolumn{2}{c}{TopK} & \multicolumn{2}{c}{BatchTopK} & \multicolumn{3}{c}{JumpReLU} \\
\cmidrule(lr){3-4}\cmidrule(lr){5-6}\cmidrule(lr){7-9}
$\log\sigma^2_{\mathrm{init}}$ & $\sigma$ & FVE & Jaccard & FVE & Jaccard & $L_0$ & FVE & Jaccard \\
\midrule
$-1.0$ & $0.607$ & $0.2517$ & $0.0750$ & $0.2459$ & $0.0758$ & $121.7$ & $0.3401$ & $0.2628$ \\
$-2.0$ & $0.368$ & $0.3511$ & $0.0862$ & $0.3599$ & $0.0826$ & $202.7$ & $0.4671$ & $0.3199$ \\
$-3.0$ & $0.223$ & $0.4708$ & $0.1094$ & $0.4833$ & $0.0968$ & $331.4$ & $0.6349$ & $0.5526$ \\
$-3.5^\ast$ & $0.174$ & & & & & $319.4$ & $0.6794$ & $0.7240$ \\
$-4.0$ & $0.135$ & $0.6107$ & $0.1686$ & $0.6232$ & $0.1311$ & $268.1$ & $0.7073$ & $0.7186$ \\
$-4.5^\ast$ & $0.105$ & & & & & $269.4$ & $0.7480$ & $0.7103$ \\
$-5.0$ & $0.082$ & $0.7200$ & $0.2624$ & $0.7366$ & $0.1881$ & $269.0$ & $0.7707$ & $0.7216$ \\
$-5.5^\ast$ & $0.064$ & & & & & $269.3$ & $0.7814$ & $0.7204$ \\
$-6.0^\ast$ & $0.050$ & & & & & $270.6$ & $0.7850$ & $0.7079$ \\
$-8.0$ & $0.050$ & $0.7494$ & $0.2919$ & $0.7692$ & $0.2101$ & $270.3$ & $0.7848$ & $0.7090$ \\
\bottomrule
\end{tabular}}
\end{table}

\mw{For both architectures FVE and Jaccard rise together, without a threshold,
across the $12\times$ range of $\sigma$. Their rank order agrees monotonically
at every point (Spearman $\rho=1$), and Pearson $r=+0.952$ (TopK) and
$+0.945$ (BatchTopK) over the six arm means. At the three noisiest points
TopK's Jaccard sits near chance ($0.075$--$0.109$) while its FVE still climbs
from $0.25$ to $0.47$. Even at the clamp floor, selection is far from stable
(Jaccard $0.292$) and FVE is $0.151$ below baseline. Compared to its own
deterministic baseline, BatchTopK recovers less than TopK at every one of the six noise scales
($80.9\%$ against $83.2\%$ at the clamp floor).}

\subsection{\texorpdfstring{\mw{Learned Thresholds}}{Learned Thresholds}}
\label{sec:jumprelu}

\mw{JumpReLU~\citep{rajamanoharan2024jumprelu} replaces the cardinality
constraint with a per-feature threshold, trained by a straight-through
estimator~\citep{bengio2013estimating} against an $L_0$ penalty. Our
implementation instead penalises the distance from a target $L_0$,
$(L_0/L_0^{\mathrm{target}}-1)^2$. Its
selection is discrete and gated, but not hard-$k$, and its noise is likewise
added before the gate. With JumpReLU, hard-$k$ selection is separated from
discrete selection in general.}

\begin{figure}[t]
\centering
\includegraphics[width=\textwidth]{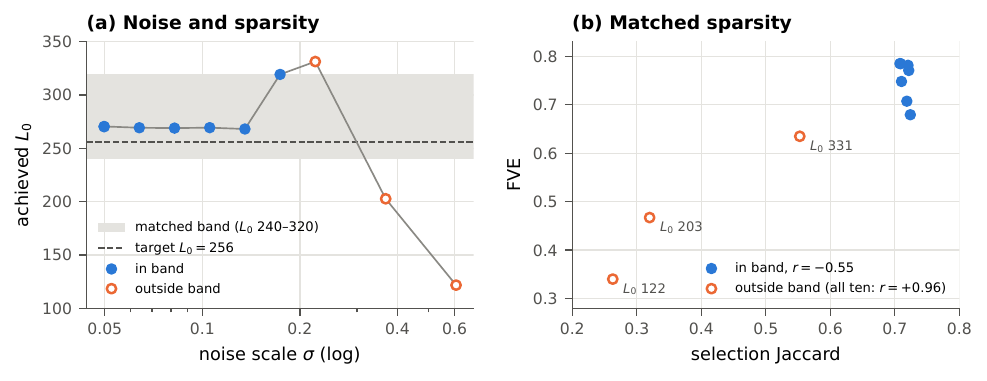}
\caption{\mw{JumpReLU under noise. (a)~Achieved $L_0$ at each noise scale
(13 seeds per point, 5 at the four added in-band points). The shaded band is
the matched-sparsity range and the dashed line the target. (b)~FVE against
selection Jaccard at the same ten points. Inside the band (filled) Jaccard
barely moves while FVE varies, and the three points outside it (open,
labelled with their $L_0$) produce the positive correlation across the whole
grid.}}
\label{fig:jumprelu}
\end{figure}

\mw{Noise moves JumpReLU's achieved sparsity itself (Table~\ref{tab:sweep},
Figure~\ref{fig:jumprelu}a). At the two noisiest points $L_0$ falls to $122$
and $203$ against the target of 256, because noise pushes pre-activations
below threshold faster than the sparsity loss can lower the threshold to
compensate. At $\sigma=0.22$ it overshoots by $30\%$. TopK and BatchTopK hold
sparsity fixed by construction, so their correlations are a clean read of
selection churn at fixed sparsity. Across JumpReLU's grid, FVE tracks $L_0$
as well ($r=+0.75$), and a 122-feature code reconstructs worse than a
270-feature one whichever features it keeps.}

\mw{To separate the two, we compare JumpReLU only where its sparsity is
roughly fixed, at grid points whose achieved $L_0$ lies between 240 and 320. After
seeing $L_0$ swing, we added four noise scales inside that band
($\log\sigma^2_{\mathrm{init}}\in\{-3.5,-4.5,-5.5,-6\}$). Across the seven
in-band points, Jaccard stays flat ($0.708$--$0.724$) while FVE rises from
$0.679$ to $0.785$ (Figure~\ref{fig:jumprelu}b). The correlation between FVE
and Jaccard is $r=+0.96$ over all ten points but $r=-0.55$ over the seven
in-band points, so the positive correlation comes entirely from the three
points where $L_0$ left the band. At matched sparsity, JumpReLU's
reconstruction does not track selection stability. What noise breaks instead
is the sparsity level the learned threshold reaches, which Top-K fixes by
construction.}

\section{\texorpdfstring{\mw{Discussion}}{Discussion}}
\label{sec:discussion}

\mw{The capacity argument
of \S\ref{sec:intro} needs the noise to limit what each latent can carry. At
fixed variance nothing is sampled, so nothing limits it, and the $L_2$
penalty the KL term reduces to does not favour general features. With every
decoder column held at unit norm, a component of size $c$ costs $c^2$ carried
by one latent but $2(c/2)^2=c^2/2$ shared equally by two latents with the same
decoder direction, so if anything the penalty favours spreading one direction
over several latents. With sampling on, the noise does limit capacity, but it
acts first on which latents enter the Top-$k$ set (\S\ref{sec:mechanism}), and
the cost we measure is reconstruction. So in neither regime does the noise
push the dictionary toward general features, as \S\ref{sec:intro}
hypothesised.}

\mw{For a stochastic sparse code, \S\ref{sec:mechanism} points to putting the noise in the
\emph{selection decision}, as Gumbel-Softmax and Concrete relaxations do for
discrete choices~\citep{jang2016categorical,maddison2016concrete} and as
perturbed optimisers do for argmax-type decisions~\citep{berthet2020learning},
by perturbing the input to the decision and differentiating through the
resulting expectation. The hard-concrete gate of \citet{louizos2018learning}
is structurally similar, with a relaxed gate deciding membership and a
separate path setting magnitude. JumpReLU's gate-then-value architecture has the same
separation without noise at the gate. A prior with mass at zero, such as the
spike-and-slab prior, is another opportunity.}

\mw{An exploratory reading across the nine fixed-variance arms with at least
six seeds each finds that arms
reconstructing better have \emph{more} near-dead features (arm-mean Spearman
$\rho=+0.90$, $p=0.0009$ and $\rho=+0.78$, $p=0.013$ at the two
thresholds).}

\mw{Some limitations of the study are that all models are trained on one small
language model, at one layer, one dictionary size and one sparsity level.
Larger scales are untested.}

\section{\texorpdfstring{\mw{Conclusion}}{Conclusion}}
\label{sec:conclusion}

\mw{At fixed variance, adding a KL term to a Top-K sparse autoencoder adds only
an $L_2$ activation penalty, and two independently built implementations of
that objective agree once the implementation details are matched. Turning
sampling on destroys reconstruction, because noise added before a hard
cardinality constraint scrambles which features are selected. Reconstruction
tracks that scrambling across a $12\times$ range of noise, for per-token and
batch-level Top-K alike. A learned threshold instead fails by losing control
of its sparsity. The vSAE therefore fails to improve on existing methods. At
fixed variance it is one of them, and with sampling on, its noise interferes
with Top-K selection.}

\section*{Acknowledgments}
We thank the Supervised Program for Alignment Research (SPAR) for facilitating
this collaboration and providing mentorship.

\bibliographystyle{plainnat}
\bibliography{references}

\appendix

\section{\texorpdfstring{\mw{SCR and TPP Disagreements}}{SCR and TPP Disagreements}}
\label{sec:e4}

\mw{This appendix asks whether a vSAE's difference from its Top-K baseline shows
up in two standard downstream interpretability metrics. It is one pair of
checkpoints with single-seed diagnostics, outside the 13-seed design, and the
vSAE trains without the AuxK loss its baseline uses, so it characterises these
two checkpoints rather than the vSAE in general.}

\paragraph{\mw{Setup.}} \mw{Two Pythia-70M SAEs (layer 3, $d=8192$), a Top-K
baseline (AuxK $\alpha=0.03125$) and its vSAE ($\beta=1$, fixed variance,
AuxK off, $1{,}474$ live features), are scored with SAEBench's spurious
correlation removal (SCR) and targeted probe perturbation (TPP)
metrics~\citep{karvonen2025saebenchcomprehensivebenchmarksparse} on
\texttt{bias\_in\_bios} (professor/nurse). SCR adapts the SHIFT
protocol~\citep{marks2025sparsefeaturecircuitsdiscovering}, which ablates the
SAE latents a probe identifies as encoding a spurious attribute and measures how
much of the accuracy gap to an unbiased probe this closes. TPP extends the idea to
multi-class disentanglement. Because the vSAE has far fewer live features, we
mask the baseline dictionary down to $N$ features by usage frequency, score
each $N$, and compare the vSAE's unmasked score against that curve at
$N=1{,}474$. A higher SCR or TPP was read as evidence of better-organised
features, but both metrics also reward selective ablation, which is easier
with fewer live features. Against an unmasked baseline the comparison carries
no evidence about organisation.}

\begin{table}[h]
\centering
\caption{\mw{SCR and TPP on the same two checkpoints, dataset and grid.}}
\label{tab:e4}
\small
\mwbox{%
\begin{tabular}{lccc}
\toprule
Metric & vSAE score & baseline curve at $N=1474$ (\texttt{top\_usage}) & margin \\
\midrule
SCR & $0.1017$ & $0.0201$ & $+0.082$ (not explained by size) \\
TPP & $0.1055$ & $0.1905$ & $-0.085$ (explained by size) \\
\bottomrule
\end{tabular}}
\end{table}

\begin{figure}[t]
\centering
\includegraphics[width=\textwidth]{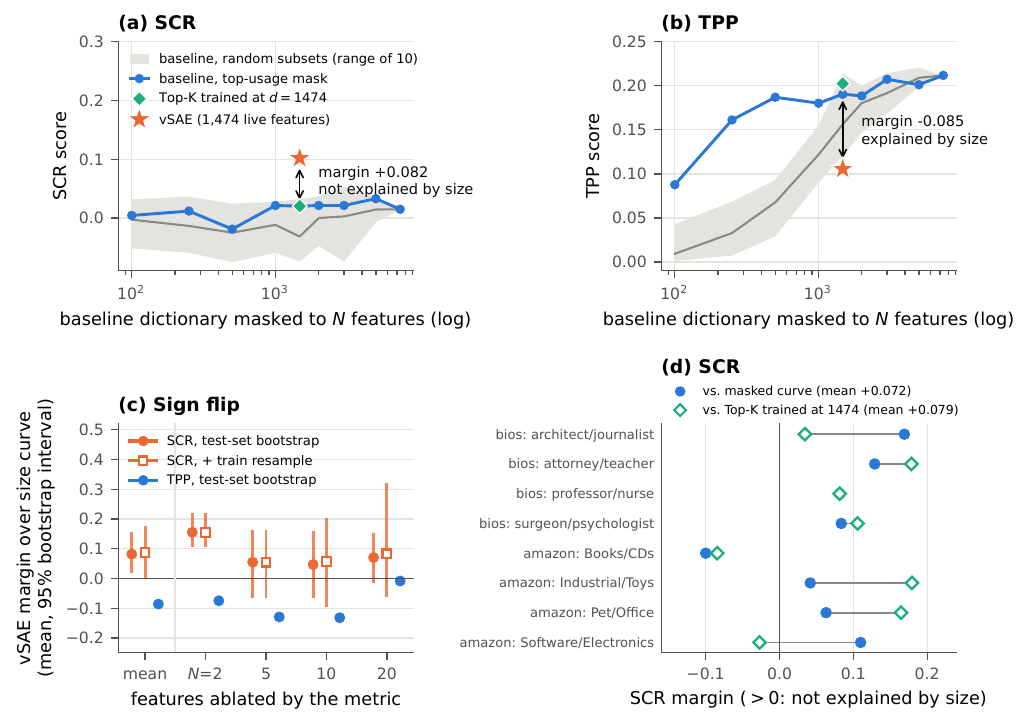}
\caption{\mw{SCR and TPP on the same two checkpoints. (a,~b)~The baseline Top-K
dictionary is masked to $N$ features by usage (blue) or at random (grey band:
range of 10 draws; grey line: their mean) and scored at each $N$; the vSAE
(star) is scored unmasked, at its 1,474 live features. The diamond is a Top-K
SAE trained from scratch at $d=1474$ (diagnostic~5). (c)~95\% bootstrap
intervals on the vSAE's margin over the top-usage curve, for the mean and at
each number of ablated features (diagnostics~1 and~2). (d)~SCR's margin at
all eight (dataset, class pair) points of diagnostic~4, against the masked
curve and against the trained-from-scratch baseline (diagnostic~5).}}
\label{fig:e4}
\end{figure}

\mw{SCR puts the vSAE above
every point of the masked curve, including the top-usage reference and a
bracket of 90 random-subset draws. Its advantage is not explained by
dictionary size. TPP puts it below both references at the same $N$
(Table~\ref{tab:e4}, Figure~\ref{fig:e4}a,b).}

\begin{enumerate}
\item \mw{\textbf{Per threshold.} Over $N\in\{2,5,10,20\}$ ablated features, SCR
  says ``not explained'' at all four and TPP ``explained'' at all four,
  decisively at $N\le10$. The disagreement is not an artifact of averaging.}
\item \mw{\textbf{Test-set noise.} A 200-draw bootstrap of the test set leaves
  both mean margins clear of zero in opposite directions, SCR at $+0.082$
  $[+0.024,+0.150]$ and TPP at $-0.086$ $[-0.090,-0.081]$. SCR's interval is
  about $15\times$ wider, and per threshold clears zero only at $N=2$.
  Re-deriving node effects from a train-set resample widens it to
  $[+0.008,+0.171]$, still clear of zero.}
\item \mw{\textbf{Different features.} We predicted that SCR's top-effect features
  would rank low in usage frequency and TPP's high. Both metrics' top features
  sit in the same $0.55$--$0.73$ usage-percentile band at every $N$, so the
  prediction fails.}
\item \mw{\textbf{Other data.} Across three more \texttt{bias\_in\_bios} pairs and
  Amazon reviews, SCR says ``not explained'' at 7 of 8 (dataset, pair) points;
  TPP says ``explained'' against its top-usage reference on both datasets,
  though its random-subset reference is a coin flip on Amazon.}
\item \mw{\textbf{The reference curve.} A Top-K SAE trained from scratch at
  $d=1474$ matches the masked curve's SCR reference to $6\times10^{-8}$ and
  scores slightly higher on TPP ($0.2025$ against $0.1905$). Across the eight
  points of item~4 it shifts SCR's reference by $0.075$ on average (up to
  $0.138$) and flips one point's sign, but the mean margin is unmoved
  ($+0.079$ against $+0.072$), 6 of 8 points still say ``not explained'', and
  TPP's two points keep their verdict.}
\end{enumerate}

\mw{One structural difference is consistent with SCR being the noisier metric.
Its per-threshold score is a ratio whose denominator
($\mathrm{clean\_acc}-\mathrm{original\_acc}$) can be small, where TPP's is a
plain difference. That does not explain the sign flip, since SCR's margin
excludes zero under both bootstraps. \citet{chanin2026reliable} have since
audited both metrics against ground truth and recommend avoiding both. At
TPP declines over training at top-$N\ge50$ and SCR at its canonical
top-$N=10$, TPP at its canonical setting does
not correlate with ground-truth quality on synthetic SAEs, and SCR ranks a
perfect oracle below 11 of 35 trained SAEs. Whichever verdict is right here,
reporting only one of the two metrics would have produced a confident claim
in either direction. Combining them would not repair this. In POPPER's benchmarks, Fisher's combined test rejected true nulls in
17--31\% of cases at a nominal 10\%~\citep{huang2025popper}.}

\section{\texorpdfstring{\mw{The Decomposition Ladder in Full}}{The Decomposition Ladder in Full}}
\label{app:ladder}

\mw{Table~\ref{tab:e1ladder} gives Cohen's $d$ of every rung of
Figure~\ref{fig:e1ladder} against the null model on all four metrics. Signs are null model minus rung, so a positive $d$ means the rung scores
lower than the null model.}

\begin{table}[h]
\centering
\caption{\mw{The decomposition ladder, as Cohen's $d$ of each rung against the
Top-K+$L_2$ null model, reading down the factors matched. $\ns$~marks
$p>0.05$ (13 seeds/arm; significant entries reach $4.30$--$5.16\sigma$).}}
\label{tab:e1ladder}
\small\setlength{\tabcolsep}{3pt}
\mwbox{%
\begin{tabular}{lccccc}
\toprule
Rung & factors matched (cumulative) & FVE ($d$) & frac.~rec.~($d$) & $<0.1\times$ ($d$) & $<0.5\times$ ($d$) \\
\midrule
R0 & KL warmup, bias form & $-5.7$ & $-8.5$ & $-2.8$ & $+13.0$ \\
R1 & + decoder init scale & $+16.5$ & $+17.7$ & $+0.8$~\ns & $+0.2$~\ns \\
R2 & + gradient projection & $+3.8$ & $+4.9$ & $-2.4$ & $-3.7$ \\
R3 & + initial weight draw & $-0.3$~\ns & $-0.8$~\ns & $-0.1$~\ns & $-0.6$~\ns \\
\bottomrule
\end{tabular}}
\end{table}

\end{document}